\pdfoutput=1

\documentclass[11pt]{article}

\usepackage[]{acl}

\usepackage{times}
\usepackage{latexsym}
\usepackage{multirow}
\usepackage{array}
\usepackage{graphicx}
\usepackage{amssymb}
\usepackage{hhline}
\usepackage{amsmath}
\usepackage{cases}
\usepackage{mathtools}
\usepackage{enumitem}
\usepackage{diagbox}
\usepackage{booktabs}
\usepackage{subfigure}

\usepackage[T1]{fontenc}
\usepackage{url}

\usepackage[utf8]{inputenc}

\usepackage{microtype}

\title{GOAL: Towards Benchmarking Few-Shot \\ Sports Game Summarization}

\newcommand{\affilsup}[1]{\rlap{\textsuperscript{\normalfont#1}}}

\author{
    Jiaan Wang\affilsup{1}~\thanks{\ \ Corresponding author.} \quad 
    ~~~ Tingyi Zhang\affilsup{1} \quad 
    ~~~ Haoxiang Shi\affilsup{2}
    \\ 
$^1$Soochow University, Suzhou, China \qquad  $^2$Waseda University, Tokyo, Japan \\
\texttt{\{jawang1,tyzhang1\}@stu.suda.edu.cn} \qquad \texttt{hollis.shi@toki.waseda.jp}
}

\begin{document}
\maketitle
\begin{abstract}
Sports game summarization aims to generate sports news based on real-time commentaries.
The task has attracted wide research attention but is still under-explored probably due to the lack of corresponding English datasets.
Therefore, in this paper, we release \texttt{GOAL}, the first English sports game summarization dataset. Specifically, there are 103 commentary-news pairs in \texttt{GOAL}, where the average lengths of commentaries and news are 2724.9 and 476.3 words, respectively.
Moreover, to support the research in the semi-supervised setting, \texttt{GOAL} additionally provides 2,160 unlabeled commentary documents.
Based on our \texttt{GOAL}, we build and evaluate several baselines, including extractive and abstractive baselines.
The experimental results show the challenges of this task still remain.
We hope our work could promote the research of sports game summarization.
%
The dataset has been released at \url{https://github.com/krystalan/goal}.

\end{abstract}

\section{Introduction}

\begin{figure}[t]
\centerline{\includegraphics[width=0.45\textwidth]{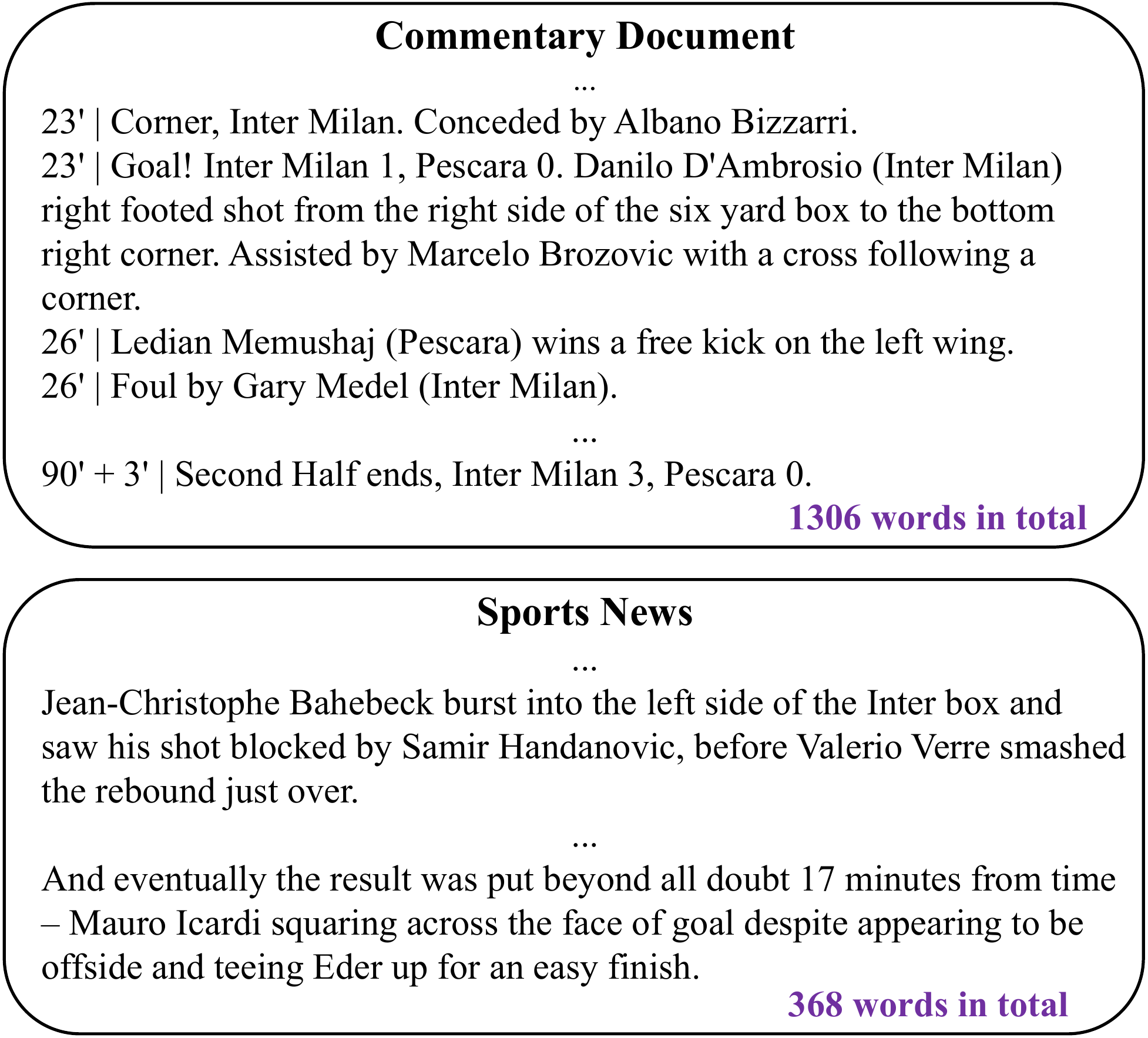}}
\caption{An \href{https://www.goal.com/en-us/match/internazionale-v-pescara/1gnkkzqjipx08k2ygwzelipyh}{example} of sports game summarization.}
\label{fig:example}
\end{figure}

Given the live commentary documents, the goal of sports game summarization is to generate the corresponding sports news~\cite{zhang-etal-2016-towards}. As shown in Figure~\ref{fig:example}, the commentary document records the commentaries of a whole game while the sports news briefly introduces the core events in the game. Both the lengthy commentaries and the different text styles between commentaries and news make the task challenging~\cite{huang-etal-2020-generating,sportssum2,ksportssum}.

\citet{zhang-etal-2016-towards} propose sports game summarization task and construct the first dataset which contains 150 samples. Later, another dataset with 900 samples is presented by~\citet{Wan2016OverviewOT}.
To construct the large-scale sports game summarization data, \citet{huang-etal-2020-generating} propose SportsSum with 5,428 samples. \citet{huang-etal-2020-generating} first adopt deep learning technologies for this task.
Further, \citet{sportssum2} find the quality of SportsSum is limited due to the original rule-based data cleaning process. Thus, they manually clean the SportsSum dataset and obtain SportsSum2.0 dataset with 5,402 samples.
\citet{ksportssum} point out the hallucination phenomenon in sports game summarization. In other words, the sports news might contain additional knowledge which does not appear in the corresponding commentary documents.
To alleviate hallucination, they propose K-SportsSum dataset which includes 7,428 samples and a knowledge corpus recording the information of thousands of sports teams and players.

Though great contributions have been made, all the above sports game summarization datasets are Chinese since the data is easy to collect. As a result, all relevant studies~\cite{zhang-etal-2016-towards,Yao2017ContentSF,Wan2016OverviewOT,Zhu2016ResearchOS,Liu2016SportsNG,lv2020generate,huang-etal-2020-generating,sportssum2} focus on Chinese, which might be difficult for global researchers to understand and study this task.
To this end, in this paper, we present \texttt{GOAL}, the first English sports game summarization dataset which contains 103 samples, each of which includes a commentary document as well as the corresponding sports news.
Specifically, we collect data from an English sports website\footnote{\url{https://www.goal.com/}\label{url:goal}} that provides both commentaries and news. We study the dataset in few-shot scenario due to: (1) When faced with real applications, few-shot scenarios are more common. Compared with the summarization datasets in the news domain which usually contain tens of thousands of samples, the largest Chinese sports game summarization dataset is still in a limited scale. (2) The English sports news is non-trivial to collect. This is because manually writing sports news is labor-intensive for professional editors~\cite{huang-etal-2020-generating,ksportssum}. We find only less than ten percent of sports games on the English website are provided with sports news.
To further support the semi-supervised research on \texttt{GOAL}, we additionally provide 2,160 unlabeled commentary documents.
The unlabeled commentary documents together with labeled samples could train sports game summarization models in a self-training manner~\cite{lee2013pseudo}.

Based on our \texttt{GOAL}, we build and evaluate several baselines of different paradigms, i.e., extractive and abstractive baselines.
The experimental results show that the sports game summarization task is still challenging to deal with.


\section{GOAL}

\subsection{Data Collection}
\href{https://www.goal.com/}{Goal.com} is a sports website with plenty of football games information. We crawl both the commentary documents and sports news from the website. 
Following~\citet{zhang-eickhoff-2021-soccer}, the football games are collected from four major soccer tournaments including the UEFA Champions League, UEFA Europa League, Premier League and Series A between 2016 and 2020.
As a result, we collect 2,263 football games in total. Among them, all games have commentary documents but only 103 games have sports news.
Next, the 103 commentary-news pairs form \texttt{GOAL} dataset with the splitting of 63/20/20 (training/validation/testing).
The remaining 2,160 (unlabeled) commentary documents are regarded as a supplement to our \texttt{GOAL}, supporting the semi-supervised setting in this dataset.
Figure~\ref{fig:example} gives an example from \texttt{GOAL}.

\begin{table}[t]
\centering
\resizebox{0.47\textwidth}{!}
{
\begin{tabular}{l|l|cc|cc}
\bottomrule[1pt]
\multicolumn{1}{c|}{\multirow{2}{*}{Dataset}}  & \multicolumn{1}{c|}{\multirow{2}{*}{\# Num.}} & \multicolumn{2}{c|}{Commentary} & \multicolumn{2}{c}{News} \\
\multicolumn{1}{c|}{}                                            & \multicolumn{1}{c|}{}                             & Avg.       & 95th pctl       & Avg.    & 95th pctl   \\ \toprule[1pt]
\multicolumn{6}{c}{Chinese}  \\ \bottomrule[1pt]
\citet{zhang-etal-2016-towards}                                                          & 150                                               & -             & -               & -          & -           \\
\citet{Wan2016OverviewOT}                                                          & 900                                               & -             & -               & -          & -           \\
SportsSum\small~\cite{huang-etal-2020-generating}                                                    & 5428                                              & 1825.6       & 3133            & 428.0     & 924         \\
SportsSum2.0\small~\cite{sportssum2}                                                     & 5402                                              & 1828.6       & -                & 406.8     & -            \\
K-SportsSum\small~\cite{ksportssum}                                                     & 7854                                              & 1200.3       & 1915            & 351.3      & 845         \\ \toprule[1pt]
\multicolumn{6}{c}{English}  \\ \bottomrule[1pt]
\texttt{GOAL} (supervised)                                              & 103                                               & 2724.9       & 3699            & 476.3     & 614         \\
\texttt{GOAL} (semi-supervised)                                       & 2160                                              & 2643.7       & 3610            & -          & -           \\ \toprule[1pt]
\end{tabular}

}
\caption{Statistics of \texttt{GOAL} and previous datasets. ``\# Num.'' indicates the number of samples in the datasets. ``Avg.'' and ``95th pctl'' denote the average and 95th percentile number of words, respectively.}
\label{table:statistics}
\end{table}

\subsection{Statistics}
As shown in Table~\ref{table:statistics}, \texttt{Goal} is more challenging than previous Chinese datasets since: (1) the number of samples in \texttt{GOAL} is significantly less than those of Chinese datasets; (2) the length of input commentaries in \texttt{GOAL} is much longer than the counterparts in Chinese datasets.

In addition, compared with Chinese commentaries, English commentaries do not provide real-time scores. Specifically, each commentary in Chinese datasets is formed as $(t,c,s)$, where $t$ is the timeline information, $c$ indicates the commentary sentence and $s$ denote current scores at time $t$.
The commentaries in \texttt{GOAL} do not provide $s$ element. Without this explicit information, models need to implicitly infer the real-time status of games.

\subsection{Benchmark Settings}
Based on our \texttt{GOAL} and previous Chinese datasets, we introduce supervised, semi-supervised, multi-lingual and cross-lingual benchmark settings. The first three settings all evaluate models on the testing set of \texttt{GOAL}, but with different training samples:
(\romannumeral1) \textit{Supervised Setting} establishes models on the 63 labeled training samples of \texttt{GOAL}; (\romannumeral2) \textit{Semi-Supervised Setting} leverages 63 labeled and 2160 unlabeled samples to train the models; (\romannumeral3) \textit{Multi-Lingual Setting} trains the models only on other Chinese datasets.

Moreover, inspired by~\citet{feng-etal-2022-msamsum,Wang2022ClidSumAB,Chen2022TheCC}, we also provide a (\romannumeral4) \textit{Cross-Lingual Setting} on \texttt{GOAL}, that lets the models generate Chinese sports news based on the English commentary documents.
To this end, we employ four native Chinese as volunteers to translate the original English sports news (in \texttt{GOAL} dataset) to Chinese. Each volunteer majors in Computer Science, and is proficient in English.
The translated results are checked by a data expert.
Eventually, the cross-lingual setting uses the 63 cross-lingual samples to train the models, and verify them on the cross-lingual testing set.

\subsection{Task Overview}
For a given live commentary document $C=\{(t_1,c_1), (t_2,c_2),...,(t_n,c_n)\}$, where $t_i$ is the timeline information, $c_i$ is the commentary sentence and $n$ indicates the total number of commentaries, sports game summarization aims to generate its sports news $R=\{r_1,r_2,...,r_m\}$. $r_i$ denotes the $i$-th news sentence. $m$ is the total number of news sentences.

\section{Experiments}

\subsection{Baselines and Metrics}
We only focus on the supervised setting, and reserve other benchmark settings for future work. The adopted baselines are listed as follows:

\begin{itemize}[leftmargin=*,topsep=0pt]
\setlength{\itemsep}{0pt}
\setlength{\parsep}{0pt}
\setlength{\parskip}{0pt}
    \item Longest: the longest $k$ commentary sentences are selected as the summaries (i.e., sports news). It is a common baseline in summarization.
    \item TextRank~\cite{mihalcea-tarau-2004-textrank} is a popular unsupervised algorithm for extractive summarization. It represents each sentence from a document as a node in an undirected graph, and then extracts sentences with high importance as the corresponding summary.
    \item PacSum~\cite{zheng-lapata-2019-sentence} enhances the TextRank algorithm with directed graph structures, and more sophisticated sentence similarity technology.
    \item PGN~\cite{see-etal-2017-get} is a LSTM-based abstractive summarization model with copy mechanism and coverage loss.
    \item LED~\cite{Beltagy2020LongformerTL} is a transformer-based generative model with a modified self-attention operation that scales linearly with the sequence length.
\end{itemize}

Among them, Longest, TextRank and PacSum are extractive summarization models which directly select commentary sentences to form sports news. PGN and LED are abstractive models that generate sports news conditioned on the given commentaries.
It is worth noting that previous sports game summarization work typically adopts the pipeline methods~\cite{huang-etal-2020-generating,sportssum2,ksportssum}, where a selector is used to select important commentary sentences, and then a rewriter conveys each commentary sentence to a news sentence.
However, due to the extremely limited scale of our \texttt{GOAL}, we do not have enough data to train the selector. Therefore, all baselines in our experiments are end-to-end methods.

To evaluate baseline models, we use \textsc{Rouge-1/2/l} scores~\cite{lin-2004-rouge} as the automatic metrics.
The employed evaluation script is based on \texttt{py-rouge} toolkit\footnote{\url{https://github.com/Diego999/py-rouge}}.

\subsection{Implementation Details}
For Longest, TextRank\footnote{\url{https://github.com/summanlp/textrank}} and PacSum\footnote{\url{https://github.com/mswellhao/PacSum}} baselines, we extract three commentary sentences to form the sports news.
$\beta$, $\lambda_1$ and $\lambda_2$ used in PacSum are set to 0.1, 0.9 and 0.1, respectively.
The sentence representation in PacSum is calculated based on TF-IDF value.
For LED baseline, we utilize \textit{led-base-16384}\footnote{\url{https://huggingface.co/allenai/led-base-16384}} with the default settings. We set the learning rate to 3e-5 and batch size to 4. We train the LED baselines with 10 epochs and 20 warmup steps.
During training, we truncate the input and output sequences to 4096 and 1024 tokens, respectively. In the test process, the beam size is 4,
minimum decoded length is 200 and maximum length is 1024.

\begin{table}[t]
\centering

\resizebox{0.45\textwidth}{!}
{
\begin{tabular}{l|ccc|ccc}
\bottomrule[1pt]
\multicolumn{1}{c|}{\multirow{2}{*}{Method}} & \multicolumn{3}{c|}{Validation} & \multicolumn{3}{c}{Testing} \\
\multicolumn{1}{c|}{}                          & R-1       & R-2     & R-L      & R-1      & R-2    & R-L     \\ \toprule[1pt]
\multicolumn{7}{c}{Extractive Methods}  \\ \bottomrule[1pt]
Longest                                       & 31.2      & 4.6     & 20.0     & 30.3     & 4.2    & 19.5    \\
TextRank                                      & 30.3      & 3.7     & 20.2     & 27.6     & 2.9    & 18.8    \\
PacSum                                        & 31.8      & 4.9     & 20.6     & 31.0     & 5.3    & 19.6    \\ \toprule[1pt]
\multicolumn{7}{c}{Abstractive Methods}  \\ \bottomrule[1pt]
PGN                                           & 34.2          & 6.3        &    22.7      &  32.8        & 5.7       & 21.4        \\
LED    & \textbf{37.8}  & \textbf{9.5}     &   \textbf{25.2}    &  \textbf{34.7}  &  \textbf{7.8}  & \textbf{24.3}         \\ \toprule[1pt]
\end{tabular}
}
\caption{Experimental results on \texttt{GOAL}.}
\label{table:results}
\end{table}

\subsection{Main Results}
Table~\ref{table:results} shows experimental results on \texttt{GOAL}. The performance of extractive baselines is limited due to the different text styles between commentaries and news (commentaries are more colloquial than new).
PGN outperforms the extractive methods since it can generate sports news not limited to original words or phrases.
However, such a LSTM-based method cannot process long documents efficiently.
LED, as an abstractive method, achieves the best performance among all baselines due to its abstractive nature and sparse attention.

\begin{figure}[t]
\centerline{\includegraphics[width=0.45\textwidth]{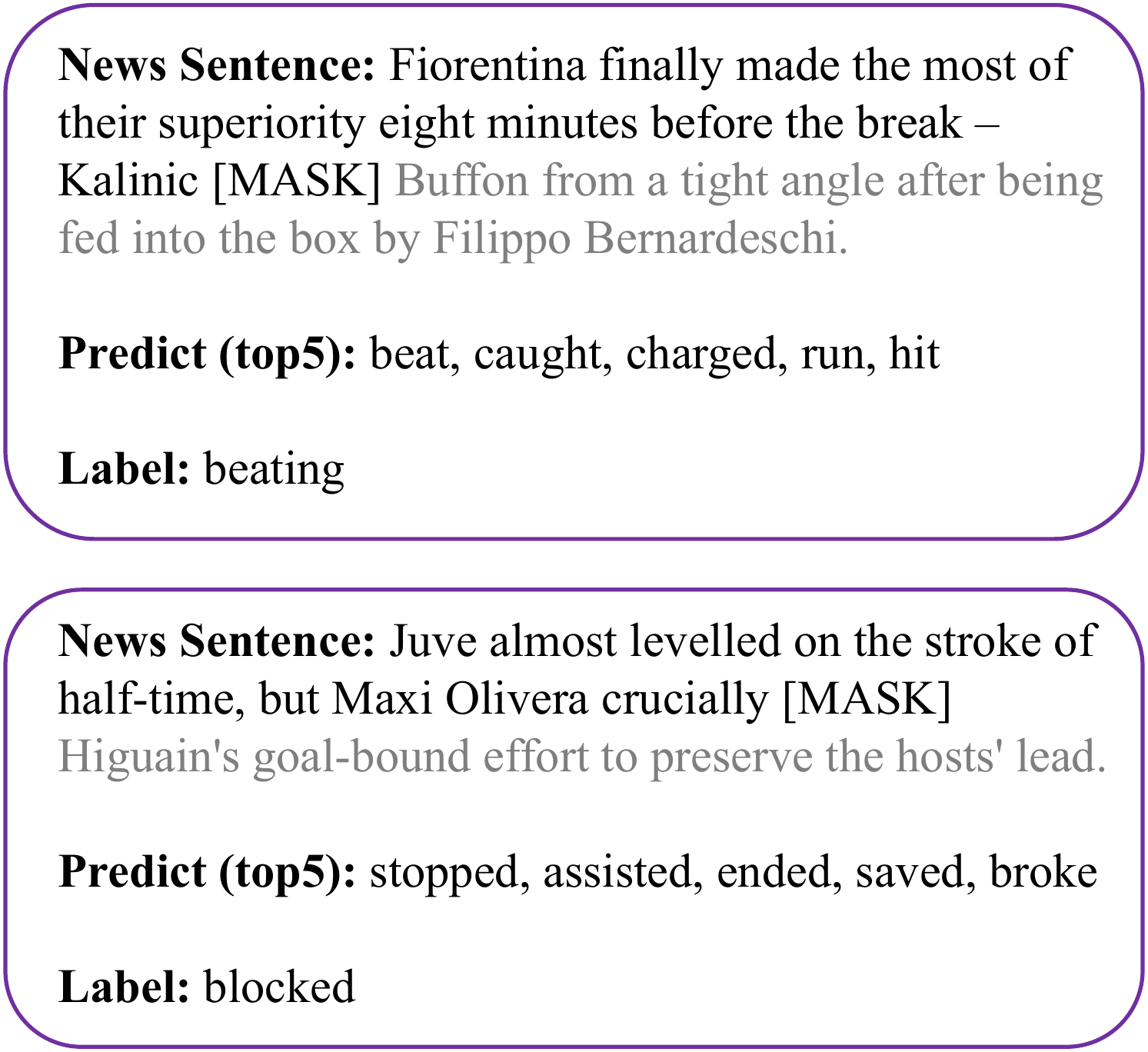}}
\caption{Predicting key verbs during generating sports news based on LED baseline. During predicting, the model is conditioned on the corresponding commentaries. The \textcolor{gray}{gray} text indicates where the model do not compute self-attention mechanism.}
\label{fig:probe}
\end{figure}

\subsection{Discussion}

Moreover, we also manually check the generated results of LED baseline, and find that the generated sports news contains many repeated phrases and sentences, and fail to capture some important events in the games.
We conjecture this is because the few-shot training samples make the baseline difficult to learn the task effectively.

To further verify whether the model is familiar with sports texts, we let the model predict the key verbs during generating the sports news, which is similar to~\citet{chen-etal-2022-probing}.
As shown in Figure~\ref{fig:probe}, when faced with a simple and common event (i.e., beating), the trained LED model could predict the right verb (i.e., beat). However, for a complex and uncommon event (i.e., blocking), the model cannot make correct predictions.

Therefore, it is an urgent need to utilize the external resources to enhance the model's abilities to know sports texts and deal with sports game summarization. For example, considering the semi-supervised setting and multi-lingual settings, where the models could make use of unlabeled commentaries and Chinese samples, respectively.
Inspired by~\citet{Wang2022ASO}, the external resources and vanilla few-shot English training samples could be used to jointly train sports game summarization models in the multi-task, knowledge-distillation or pre-training framework.
In addition, following~\citet{Feng2021ASO}, another promising way is to adopt other long document summarization resources to build multi-domain or cross-domain models with sports game summarization.

\section{Conclusion}
In this paper, we present \texttt{GOAL}, the first English sports game summarization dataset, which contains 103 commentary-news samples. Several extractive and abstractive baselines are built and evaluated on \texttt{GOAL} to benchmark the dataset in different settings.
We further analyze the model outputs and show the challenges still remain in sports game summarization.
In the future, we would like to (1) explore the semi-supervised and multi-lingual settings on \texttt{GOAL}; (2) leverage graph structure to model the commentary information, and generate sports news in a graph-to-text manner~\cite{ijcai2021-524}, or even consider the temporal information (i.e., $t_i$) in the graph structure~\cite{10.1007/978-3-031-00129-1_10}.


\bibliography{anthology}




\end{document}